\documentclass[runningheads,orivec]{llncs}
\usepackage[T1]{fontenc}
\usepackage{graphicx}
\usepackage{booktabs}
\usepackage[misc]{ifsym}
\newcommand{\corr}{(\Letter)}

\usepackage[nocompress]{cite}
\usepackage{amsmath}
\usepackage{amssymb}
\usepackage{multirow}
\usepackage{microtype}
\usepackage{xcolor}
\usepackage[hidelinks]{hyperref}

\newcommand{\llbracket}{[\![}
\newcommand{\rrbracket}{]\!]}

\begin{document}

\title{CQD-SHAP: Explainable Complex Query Answering via Shapley Values}

\toctitle{CQD-SHAP: Explainable Complex Query Answering via Shapley Values}

\author{Parsa Abbasi\orcidID{0009-0002-4824-3645} \corr \and
Stefan Heindorf\orcidID{0000-0002-4525-6865}}

\tocauthor{Parsa Abbasi, Stefan Heindorf}

\authorrunning{P. Abbasi and S. Heindorf}

\institute{Department of Computer Science, Paderborn University, Germany\\
\email{\{parsa.abbasi, heindorf\}@uni-paderborn.de}}

\maketitle

\begin{abstract}
Complex query answering (CQA) goes beyond the widely studied link prediction task by addressing more sophisticated queries that require multi-hop reasoning over incomplete knowledge graphs (KGs). Research on neural and neurosymbolic CQA methods is still an emerging field. Almost all of these methods can be regarded as black-box models, which may raise concerns about user trust. Although neurosymbolic approaches like CQD are slightly more interpretable, allowing intermediate results to be tracked, the importance of different parts of the query remains unexplained. In this paper, we propose CQD-SHAP, a novel framework that computes the contribution of each query part to the ranking of a specific answer. This contribution explains the value of leveraging a neural predictor that can infer new knowledge from an incomplete KG, rather than a symbolic approach relying solely on existing facts in the KG. CQD-SHAP is formulated based on Shapley values from cooperative game theory and satisfies all fundamental Shapley axioms. Automated evaluation of these explanations in terms of necessary and sufficient explanations, and comparisons with various baselines, show the consistent effectiveness of this approach across all studied datasets and query types.

\end{abstract}

\section{Introduction}
Knowledge graphs (KGs) are a rich resource for retrieving fact-based answers to queries. However, real-world KGs are often incomplete. For example, in Freebase~\cite{Bollacker2008Freebase}, among more than 3 million persons, only 25\% have a recorded nationality and only 6\% have information about their parents~\cite{West2014Knowledge}. Similarly, in Wikidata~\cite{Vrandecic2014Wikidata} only 50\% of the artists have a recorded place of birth~\cite{Zhang2022Enriching}. Consequently, retrieving knowledge solely by traversing the graph, a procedure we refer to as \emph{symbolic execution}, is prone to missing relevant results.

Although there has been extensive research on the link prediction task~\cite{Wang2021Survey}, these methods focus on answering simple one-hop queries, i.e., inferring a missing link between pairs of entities. In practice, however, queries often combine conditions through logical operators (e.g., conjunction or disjunction) and require multi-hop reasoning. Following established terminology in prior work, we refer to such queries as \emph{complex queries}~\cite{Ren2024Neural}. To answer such queries on incomplete KGs, a number of neural and neurosymbolic approaches have recently been proposed~\cite{Ren2024Neural}. One of the first and most influential among them is CQD~\cite{Arakelyan2021CQD}, a neurosymbolic model, which decomposes a query into atomic link predictions, hereafter called \emph{neural execution}, and combines the results through fuzzy logic.

Despite their success, neural query answering methods often lack transparency, and it is not always clear why a particular answer is produced. While some explanation methods have been proposed for one-hop link prediction models~\cite{Rossi2022Explaining,Chang2024Path,Mohapatra2023Integrating,Zhang2023PaGE-Link}, they are not directly applicable to complex queries and typically focus on identifying \emph{important triples} in the KG for a given prediction. In contrast, complex queries may require different types of evidence across different parts of the query, calling for novel explanation definitions. In this work, we focus on the constituent parts of a query, which we refer to as \emph{atoms}, and aim to explain the \emph{importance of each atom} contributing to the ranking of an answer.

We argue that quantifying the contribution of query atoms is useful in different ways:
(1) it helps domain experts identify and correct gaps in the KG by revealing which parts of a query rely most strongly on inferred knowledge;
(2) it supports users and domain experts in debugging and refining queries by showing which query atoms have the greatest influence on the final ranking of an answer, and
(3) it helps data scientists to debug and improve complex query answering models. 

The practical relevance of atom-level explanations is best demonstrated by example. Consider the query: ``Which drugs are prescribed for diabetes \emph{and} cause kidney toxicity?'' A neurosymbolic reasoner such as CQD scores drugs prescribed for diabetes and drugs causing kidney toxicity, then combines the resulting scores via a \emph{t-norm} (e.g., multiplication) to rank answer entities. Suppose \texttt{Insulin} appears among the top-ranked predictions. Our explanation approach can help the user understand why the neurosymbolic reasoner produces this ranking. Specifically, it reveals whether the ranking is driven primarily by the fact that \texttt{Insulin} is prescribed for diabetes in the KG, or by the (incorrect) association that it causes kidney toxicity. For instance, the method might assign a contribution score of +10 to the first atom (``prescribed for diabetes'') and +450 to the second (``causes kidney toxicity''), indicating that the latter contributed far more to achieving the high rank. Such explanations enable users to critically examine that part of the query and its associated subgraph, in order to assess whether the system is relying on misleading correlations, behaving incorrectly, or potentially uncovering a valid but previously overlooked insight.

In this paper, we propose \emph{CQD-SHAP}, a novel approach to explaining the results of the CQD model using \emph{Shapley values}~\cite{Shapley1953Value}, which are grounded in cooperative game theory. We define a Shapley game over query atoms, treating each atom as a player. For every subset of atoms (coalition), the game’s value is defined as the rank of the target entity when atoms in the coalition are executed using the neural model and the remaining atoms are executed symbolically. The Shapley value of an atom quantifies the \emph{average contribution} of its neural execution to the ranking of the target entity across all coalitions.

For instance, in the previous example, if the Shapley value of the second atom (``causes kidney toxicity'') is 450, this indicates that, averaged over all possible execution choices of the other atom, executing this atom neurally improves the ranking of the answer \texttt{Insulin} by 450 positions. Shapley values are supported by a mathematically rigorous foundation, and any well-defined Shapley game inherits useful formal properties. One such property is \emph{efficiency}, which provides additional value for explainability. In our framework, efficiency means that the sum of Shapley values across all atoms in a query equals the overall improvement achieved by executing the entire query neurally, compared to executing it symbolically (i.e., by traversing the KG alone).
We summarize our main contributions as follows:

\begin{itemize}
    \item To the best of our knowledge, this is the first approach to explaining the results of neural complex query answering models using Shapley values.
    \item We propose a formal definition of explanation for the complex query answering task at the level of query atoms.
    \item We introduce \emph{CQD-SHAP}, which quantifies the contribution of using a neural model for each query atom to the ranking of a target answer.
    \item Through comprehensive experiments on FB15k-237, NELL995, and their harder variants (FB15k-237+H, NELL995+H), we assess \emph{CQD-SHAP} explanations via \emph{necessity} and \emph{sufficiency} criteria, demonstrating consistent interpretability and meaningfulness across diverse query difficulties.
\end{itemize}

All resources used in this work are publicly available in our GitHub repository.\footnote{\url{https://github.com/ds-jrg/CQD-SHAP}}

\section{Related Work}
\paragraph{Neural Query Answering.} Unlike symbolic query answering methods, such as SPARQL, which can only retrieve incomplete sets of answers directly reachable within the observed graph, neural methods attempt to infer missing information using neural networks, thereby recovering additional \textit{hard answers} that symbolic methods cannot attain.
Following the categorization suggested by Ren et al.~\cite{Ren2024Neural}, \emph{Neural Query Answering} methods execute all parts of a query in the latent space. Some neural approaches, such as GQE~\cite{Hamilton2018Embedding}, embed queries, entities, and relations as vectors and model logical operations as geometric transformations. However, other works, such as MPQE~\cite{Daza2020Message}, do not employ any explicit execution of logical operations and instead process the entire query graph simultaneously.

\paragraph{Neurosymbolic Query Answering.} These methods lie between neural and symbolic methodologies, executing query atoms (projections) neurally while simulating logical operations when combining results, enabling greater interpretability. Such operations can be mimicked in different ways; for example, Query2Box~\cite{Ren2020Query2box} leverages geometric operations, whereas CQD~\cite{Arakelyan2021CQD} and its variants~\cite{Demir2023LitCQD,Arakelyan2023Adapting,Ott2025Training,Wang2023Logical} employ fuzzy logic. In this work, we focus on neurosymbolic approaches and specifically adopt the design principles underlying CQD.

\paragraph{Explainable Link Prediction.} CQA methods, and in particular neurosymbolic approaches such as CQD~\cite{Arakelyan2021CQD}, can be viewed as extensions of the well-studied link prediction problem. Recent advances in Knowledge Graph Embedding (KGE) have yielded strong performance across graph-based tasks, including link prediction, but their black-box nature makes predictions difficult to interpret. Since many real-world applications require both high performance and transparency, various explanation methods have been proposed for link prediction. For example, Kelpie~\cite{Rossi2022Explaining} explains why a given tail entity $o$ is top-ranked for a query $(s,p,?)$ by identifying training facts that are \emph{necessary} or \emph{sufficient} for the prediction. CRIAGE~\cite{Pezeshkpour2019Investigating} identifies influential facts by applying adversarial perturbations to the KG and observing their impact on the prediction. Power-Link~\cite{Chang2024Path} and PaGE-Link~\cite{Zhang2023PaGE-Link}, instead focus on path-based explanations, finding influential paths whose intermediate nodes and edges strongly impact a predicted link. However, all these methods primarily address single-projection link prediction, whereas complex queries require multiple projections whose contributions may interact to produce the final answer. In our work, we address complex queries that require multiple projections, where the results of individual projections may depend on each other to produce the final answer. We do not assume that every projection is executed by a link prediction model; instead, we measure the importance of each projection by assessing how much replacing its neural execution via a link prediction model with a symbolic execution (i.e., a simple graph look-up) affects the final result. Our explanation approach considers different paths leading to the final answer set, which makes it partially related to path-based explainable link prediction methods. Moreover, since our work builds on CQD, the path to each final answer remains explicit and accessible to the end user.

\paragraph{Explainable Query Answering.} To the best of our knowledge, no prior work has directly attempted to explain complex query answering over KGs. However, there is considerable research on query explanation in the context of traditional databases. Gathani et al. \cite{Gathani2020Debugging} surveys various techniques for explaining and debugging database queries, and our approach relates to their ``Why and Why Not''~\cite{Chapman2009Why} strategy. Similar to our study, some works such as~\cite{Deutch2022Computing} and~\cite{Livshits2021Shapley} adopt Shapley values for explanation. However, they quantify the contribution of database facts or tuples to query results rather than explaining the query itself. Furthermore, both methods focus on aggregation or Boolean queries, which have numerical outputs such as ``Is there a direct or one-stop flight from a German airport to an Iranian airport?''---where the answer is simply Yes~(1) or No~(0). Beyond databases, Bienvenu et al.~\cite{Bienvenu2024Shapley} addresses knowledge bases, but again focuses on quantifying fact contributions rather than explaining the query.

\paragraph{Shapley Values for Ranking and Retrieval.}
Although there is a lack of research on using Shapley values to explain complex queries, this problem can be approached through the lens of ranking and retrieval, as both domains output an ordered list of entities ranked by relevance. Shapley values, and particularly SHAP~\cite{Lundberg2017Unified}, were originally introduced for models that output a single score. Therefore, applying them to explain the outcome of a ranking model is not straightforward and requires a novel definition of a value function capable of producing a single score from a ranked list. In recent years, this challenge has been the focus of several studies. For instance, Heuss et al. \cite{Heuss2025RankingSHAP} introduced \textit{RankingSHAP}, which generates the feature contributions for ordered lists, but it does not satisfy some of the fundamental properties of Shapley values, such as \textit{efficiency} and \textit{monotonicity}, as correctly demonstrated in \cite{Chowdhury2025RankSHAP}. Chowdhury et al. \cite{Chowdhury2025RankSHAP} proposed \textit{RankSHAP} by introducing a generalized ranking value function and discussing the properties required to satisfy the four Shapley value axioms in ranking. However, both methodologies focus on explaining feature contributions for the entire ranking. In contrast, Pliatsika et al. \cite{Pliatsika2025ShaRP} focus on explaining a single ranked outcome and propose a different method for quantifying the payoff with respect to the ranking of a particular target entity, referred to as the \emph{Quantity of Interest (QoI)}. Furthermore, they demonstrate that these proposed QoIs satisfy all the fundamental axioms.
We follow this approach and adopt the same QoI definitions.

\section{Background}

We briefly introduce the task of complex query answering and fundamentals of Shapley values, before introducing our approach in the next section.

\paragraph{Knowledge Graph.}
A triple-based knowledge graph $\mathcal{G} \subseteq \mathcal{E} \times \mathcal{R} \times \mathcal{E}$ comprises a set of triples $\langle s, p, o \rangle$, where each triple represents a factual statement involving a relation $p \in \mathcal{R}$ between the subject entity $s \in \mathcal{E}$ and the object entity $o \in \mathcal{E}$. Here, $\mathcal{E}$ and $\mathcal{R}$ denote the sets of entities and relations, respectively.

\paragraph{Queries.}
Following Arakelyan et al.~\cite{Arakelyan2021CQD}, queries over a KG can be expressed in first-order logic (FOL). Each triple is represented as an atomic formula $a = p(s, o)$. By combining these atomic formulas using logical operators such as conjunction ($\wedge$) and existential quantification ($\exists$), one can construct a class of FOL queries known as \emph{conjunctive queries}. In line with their work, we focus on \emph{existential positive first-order} (EPFO) queries, which additionally allow disjunction ($\vee$), and transform them into disjunctive normal form (DNF). The formal definition of a DNF query $\mathcal{Q}$ is given in Equation~\eqref{eq:dnf_query}. In this formulation, $V_1, \ldots, V_m$ denote variables bound to nodes via existential quantification, and each $e \in \mathcal{E}$ represents a fixed input entity, also called an \emph{anchor} node. The variable $V_A$ is the target of the query, and we denote the answer set of $\mathcal{Q}$ by $\mathcal{E}_A = \llbracket \mathcal{Q}[V_A] \rrbracket$.

\begin{align}
\mathcal{Q}[V_A]
    &\triangleq\ ?V_A: \exists\, V_1,\ldots,V_m\ \notag
     \Big[\, (a_1^1 \wedge \cdots \wedge a_{n_1}^1)\ \vee\ \cdots \vee\ (a_1^d \wedge \cdots \wedge a_{n_d}^d) \,\Big] \notag \\
    &\text{where} \quad
    a_i^j =
    \begin{cases}
      p(e, V), & 
        \begin{array}{l}
          V \in \{V_A, V_1, \ldots, V_m\}, \\
          e \in \mathcal{E},~p \in \mathcal{R}
        \end{array} \\[4ex]
      p(V, V'), &
        \begin{array}{l}
          V \in \{V_A, V_1, \ldots, V_m\}, \\
          V' \in \{V_A, V_1, \ldots, V_m\}, \\
          V \neq V',~p \in \mathcal{R}
        \end{array}
    \end{cases}
    \label{eq:dnf_query}
\end{align}

\paragraph{Symbolic Query Answering.} In a symbolic approach, the query $\mathcal{Q}$ is answered by binding $V_A$ to each entity in the KG and determining for which entities $\mathcal{Q}[V_A]$ holds true. This approach implicitly assumes that the KG is complete. 

\paragraph{Neurosymbolic Query Answering.}
In practice, KGs are often incomplete and so is the result of a symbolic approach. Neurosymbolic CQA models~\cite{Ren2024Neural} address this limitation by inferring missing information and producing approximate answer sets. For evaluation, we assume access to a complete knowledge graph $G_{\mathrm{test}}$, from which we randomly sample a subset of triples to construct an incomplete graph $G_{\mathrm{valid}} \subseteq G_{\mathrm{test}}$. We train a neurosymbolic CQA model on $G_{\mathrm{valid}}$, use it to answer queries, and compare its predictions to ground-truth answers obtained by executing a symbolic approach on $G_{\mathrm{test}}$. The goal of the neurosymbolic model is to approximate the answers that a symbolic method would return on the complete graph, despite only having access to the incomplete graph during training.

\paragraph{Complex Query Decomposition (CQD).}
The goal of CQD~\cite{Arakelyan2021CQD} is to find a mapping from variables to entities that maximizes the score of $\mathcal{Q}$, as shown in Equation~\eqref{eq:dnf_argmax}. For this, CQD computes a prediction score $\rho(\mathbf{e}_s, \mathbf{e}_p, \mathbf{e}_o)$ for each atom in the query, based on the embedding vectors of its subject, predicate, and object, $\mathbf{e}_s, \mathbf{e}_p, \mathbf{e}_o \in \mathbb{R}^{\ell}$, where $\ell$ denotes the embedding dimension. These scores are then aggregated using continuous generalizations of logical operators, such as t-norms ($\top$) for conjunctions and t-conorms ($\perp$) for disjunctions.

\begin{align}
	&\underset{V_A, V_1, \ldots, V_m \in \mathcal{E}}{\arg \max }
	\Big[\, (a_1^1 \top \cdots \top a_{n_1}^1)\ \perp \cdots
	\perp (a_1^d \top \cdots \top a_{n_d}^d)\,\Big]
	\notag\\
	&\hspace{0.8em} \text{where}~ 
	a_i^j = 
	\begin{cases}
		\rho(\mathbf{e}_e, \mathbf{e}_p, \mathbf{e}_V), &
		\begin{array}{@{}l@{}}
			V \in \{V_A, V_1, \ldots, V_m\}, \\
			e \in \mathcal{E},~p \in \mathcal{R}
		\end{array} \\[3ex]
		\rho(\mathbf{e}_V, \mathbf{e}_p, \mathbf{e}_{V'}), &
		\begin{array}{@{}l@{}}
			V \in \{V_A, V_1, \ldots, V_m\}, \\
			V' \in \{V_A, V_1, \ldots, V_m\}, \\
			V \neq V',~p \in \mathcal{R}
		\end{array}
	\end{cases}
	\label{eq:dnf_argmax}
\end{align}

Let $z(e_i)=(a_1^1 \top \cdots \top a_{n_1}^1)\ \perp \cdots \perp (a_1^d \top \cdots \top a_{n_d}^d)$ denote the score assigned by the neurosymbolic CQA model to entity $e_i \in \mathcal{E}_A$ as a candidate answer to the query $\mathcal{Q}$. To compute this score, we substitute $V_A \leftarrow e_i$ and replace the remaining variables with the entities that maximize the final score.

Then, $z(e_i)$ estimates the likelihood that the entity $e_i$ is an answer to $\mathcal{Q}$ in the complete KG. In practice, the model ranks all candidate entities in the graph (or a subset thereof) according to their prediction scores, producing an answer set ${\mathcal{E}_A} \subseteq \mathcal{E}$. 
Therefore, for any $e_i, e_j \in {\mathcal{E}_A}$, if the rank $r_i < r_j$, then $e_i$ is more likely to be an answer to the query than $e_j$.

\paragraph{Shapley Values.}
Shapley values, originally introduced by \textit{Lloyd Shapley} in cooperative game theory~\cite{Shapley1953Value}, are widely used in Explainable AI (XAI) and popularized by SHAP~\cite{Lundberg2017Unified}. They provide a fair and principled way to attribute a model's output to its input features. Formally, let \( P = \{1, 2, \dots, p\} \) be a set of \textit{players}, and let \( val: 2^P \rightarrow \mathbb{R} \) be a \textit{value function} that assigns a real number to each subset \( S \subseteq P \), with \( val(\{\}) = 0 \). The \textit{Shapley value} \( \phi_i(val) \) for a player \( i \in P \) is defined as:
\begin{equation}
\phi_i(val) := \sum_{S \subseteq P \setminus \{i\}} \frac{|S|! \, (|P| - |S| - 1)!}{|P|!} \, \left[ val(S \cup \{i\}) - val(S) \right]
\label{eq:shapley}
\end{equation}
This formula computes the average marginal contribution of player~\( i\) across all possible subsets \( S \subseteq P \setminus \{i\} \), weighted by the number of permutations in which \( i \) joins the coalition after \( S \)---ensuring the properties of efficiency, symmetry, linearity, and null player~\cite{Winter2002Shapley}.

\section{Explainable Complex Query Answering} \label{methodology}

\begin{figure}[t]
    \centering
    \includegraphics[width=1\linewidth]{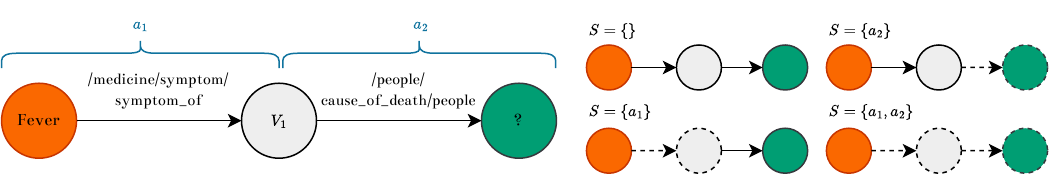}
    \caption{Illustration of all possible partial queries $\mathcal{Q}_S$ for a query consisting of two sequential projections ($2p$). We execute the atom $a_i$ using the symbolic approach when $a_i \notin S$ (solid lines), and using the neural approach when $a_i \in S$ (dashed lines).}
    \label{fig:2p}
\end{figure}

Given a query $\mathcal{Q}$ which consists of the atoms $\mathcal{A} = \{a_1, \ldots, a_{|\mathcal{A}|}\}$, each atom can either be answered with a symbolic approach (possibly returning an incomplete answer set) or with a neural approach. Fig.~\ref{fig:2p} provides an illustration of this process, showing all possible partial queries for a simple case of two sequential projections ($2p$). Each path represents a different combination of neural and symbolic executions, with dashed and solid lines indicating the two cases, respectively.
Finally, the answer is composed of the intermediate answers to the query atoms.
Our goal is to use Shapley values to quantify the importance of employing a neural approach, rather than a symbolic one, for each atom when ranking a specific answer entity. Given a query $\mathcal{Q}$ and entity $e_i \in \mathcal{E}_A$, we define the cooperative game as follows:
\begin{itemize}
\item The players are the atoms: $P:=\mathcal{A}$
\item For any coalition $S \subseteq P$, the partial query $\mathcal{Q}_S$ is defined such that atoms in $S$ are answered neurally and the remaining atoms symbolically
\item The value function $val_{e_i}({\mathcal{Q}_S})$ evaluates how good the answer $e_i$ is to the query $\mathcal{Q}_S$
\end{itemize}
Using this definition of a game, we obtain a Shapley value for each atom of the query with respect to a given target entity.
In the following, we define the execution of a partial query $\mathcal{Q}_S$ and the value function $val$ in more detail. 

For atoms in $S$, execution is performed using a neural approach (e.g., the link predictor in CQD~\cite{Arakelyan2021CQD}) which enables inference over incomplete KGs. For atoms not included in $S$, the answer is computed in a purely symbolic manner, relying solely on the observed facts in the KG.
\begin{align}
	&\mathcal{Q}_S[V_A] \triangleq \notag \underset{V_A, V_1, \ldots, V_m \in \mathcal{E}}{\arg \max }
	\Big[\, (a_1^1 \top \cdots \top a_{n_1}^1)\ \perp \cdots
	\perp (a_1^d \top \cdots \top a_{n_d}^d)\,\Big]
	\notag\\
	&\hspace{1.5em} \text{where}~ 
	a_i^j = 
	\begin{cases}
		\begin{array}{l}
            \rho(\mathbf{e}_e, \mathbf{e}_p, \mathbf{e}_V) \quad \text{or} \quad
             \rho(\mathbf{e}_V, \mathbf{e}_p, \mathbf{e}_{V'}),
            \end{array}
            & \text{if } a_i^j \in S \\[3ex]
            \begin{array}{l}
            p(e, V) \quad \text{or} \quad
            p(V, V'),
            \end{array}
            & \text{otherwise}
	\end{cases}
	\label{eq:query_s}
\end{align}

To execute such a query, we employ the combinatorial optimization introduced in CQD. Query execution begins with atoms that contain an anchor, i.e., $p(e, V)$ or $\rho(\mathbf{e}_e, \mathbf{e}_p, \mathbf{e}_V)$. If the neural approach is selected ($\rho(\mathbf{e}_e, \mathbf{e}_p, \mathbf{e}_V)$), the neural link predictor assigns a score to each entity, representing the likelihood that the entity is connected to $e$ via relation $p$. As in the original CQD approach, if the atom involves the target variable ($V_A$), the model produces a score for every entity. If, instead, the atom corresponds to an intermediate step in the beam search, the entities are sorted in descending order by score, and the top-$k$ are returned.

If the symbolic approach is chosen (i.e., $p(e, V)$), we identify all entities connected to $e$ via relation $p$ in the observed graph. By \emph{observed graph}, we mean the validation graph when explaining a test query, and the training graph when explaining a validation query. These entities are assigned a constant score (in our case, 1), while all other entities receive a low score close to 0. If this atom does not contain the target variable $V_A$, the entities are sorted in descending order, and the top-$k$ candidates are returned; otherwise all results are returned.

This formulation ensures that the importance assigned to each atom reflects both its direct logical contribution and the impact of the specific reasoning mechanism (neural or symbolic) by which the query is executed.

Now that the definition of the game is in hand, the payoff or the value function of a coalition~$S$ should be defined. Following Pliatsika et al. \cite{Pliatsika2025ShaRP}, we can define different quantities of interest (QoIs) as the outcome of the value function based on the position of answer~$e_i$ in the ranking. In this work, we employ the function \emph{$\Delta$Rank QoI} as defined in the following.
Suppose $\mathcal{E}_A$ denotes the answer set for the partial query~$\mathcal{Q}_S$. In addition, let $\mathcal{E}_{\mathrm{easy}}$ denote the set of \emph{easy answers} that can be obtained through symbolic execution on the observed graph, and $\mathcal{E}_{\mathrm{hard}}$ the set of \emph{hard answers} that cannot \cite{Ren2024Neural,Ren2020Query2box}.
Following CQD~\cite{Arakelyan2021CQD}, we adopt the \emph{filtered} setting~\cite{Bordes2013Translating} for computing rankings in order to ensure a fair evaluation. Let $\mathcal{E}_A^{(i)}$ denote the set of candidate answers obtained by excluding all easy answers and all hard answers except the current target hard answer $e_i$:
\begin{align}
\mathcal{E}_A^{(i)}
= \mathcal{E}_A \setminus \Big( \mathcal{E}_{\mathrm{easy}} 
   \cup \big(\mathcal{E}_{\mathrm{hard}} \setminus \{e_i\}\big)\Big).
\label{eq:filteredlist}
\end{align}

The rank $r_i$ of an entity $e_i$ is defined as its position in the filtered, ordered list obtained by sorting the entities $\mathcal{E}_A^{(i)}$ in descending order of their score: \begin{align}
r_i=|\{e_j \in \mathcal{E}_A^{(i)} \backslash\{e_i\}: z(e_j)>z(e_i) \}| + 1\label{eq:rankeq}
\end{align}

The \emph{$\Delta Rank$ QoI} is the difference between the rank of $e_i$ in the answer to the partial query $\mathcal{Q}_S$ and its rank in the result of the complete symbolic query $\mathcal{Q}_{\{\}}$, as formalized in Equation~\eqref{eq:rankqoi}. This formulation ensures that the value of the empty coalition, $val_{e_i}(\mathcal{Q}_{\{\}})$, is equal to zero, as required in the Shapley framework, thereby guaranteeing that the fundamental axioms are satisfied.
\begin{align}
val_{e_i}(\mathcal{Q}_S) = r_i(\mathcal{Q}_{\{\}}) - r_i(\mathcal{Q}_S) \label{eq:rankqoi}
\end{align}

Finally, to formally measure the marginal contribution of each atom to the ranking of a specific entity $e_i$, we compute the Shapley value for each atom $a \in \mathcal{A}$ by plugging our definition of a cooperative game into the original Shapley definition, given in Equation~\eqref{eq:shapley}. The Shapley value then aggregates the expected contribution of atom $a$ over all possible coalitions $S \subseteq \mathcal{A} \setminus \{a\}$:
\begin{align}
\phi_{a}(e_i)
    &:= \sum_{S \subseteq \mathcal{A} \setminus \{a\}}
        \frac{|S|!~(|\mathcal{A}| - |S| - 1)!}{|\mathcal{A}|!}\,
        \Delta val_{e_i}(S, a) 
        \label{eq:phi_def}
\end{align}
where $\Delta val_{e_i}(S, a) = val_{e_i}(\mathcal{Q}_{S \cup \{a\}}) - val_{e_i}(\mathcal{Q}_S)$, $|\mathcal{A}|$ denotes the total number of atoms in the query, and $\Delta val_{e_i}(S, a)$ represents the marginal change in the outcome when the atom $a$ is added to the coalition $S$.

Our defined value functions align with the metrics proposed in Chowdhury et al. \cite{Chowdhury2025RankSHAP}. For example, for the $\Delta$Rank QoI, we can set the gain function to the linear form, $g(\text{rel}_j) = \text{rel}_j$, and use no discounting, $h(j) = 1$. In this case, the relevance score $\text{rel}_j$ can be defined as the difference between the rank of $e_j$ in $\mathcal{Q}_S$ and its rank in $\mathcal{Q}_{\{\}}$, ensuring that the relevance sensitivity property holds. As proven in that study, such a function satisfies all the main Shapley value axioms. Moreover, since the number of atoms in a complex query is relatively small, no approximation is required, and the exact Shapley values can be computed.

Building upon the Shapley value’s \emph{efficiency} axiom, one can derive that the sum of the Shapley values for all atoms (a total of $|\mathcal{A}|$ atoms) with respect to a given answer $e_i$ is equal to the value of the grand coalition, i.e., when all atoms are executed neurally ($\mathcal{Q}_{\{a_1, \ldots, a_{|\mathcal{A}|}\}}$). This can be expressed as:
\begin{align}
\phi(e_i) = \sum_{a \in \mathcal{A}}\phi_{a}(e_i) &= val_{e_i}(\mathcal{Q}_{{a_1, \ldots, a_{|\mathcal{A}|}}}) = r_i(\mathcal{Q}_{\{\}}) - r_i(\mathcal{Q}_{{a_1, \ldots, a_{|\mathcal{A}|}}})
\label{eq:efficiency}
\end{align}

In other words, the sum of the Shapley values over all atoms equals the difference in the rank of entity $e_i$ between fully neurosymbolic query execution ($\mathcal{Q}_{\{a_1, \ldots, a_{|\mathcal{A}|}\}}$) and purely symbolic execution ($\mathcal{Q}_{\{\}}$).

\section{Evaluation}

After describing our evaluation setup, including performance metrics, evaluation scenarios and baselines (Section~\ref{subsec:evaluation-setup}), we provide quantitative results (Section~\ref{subsec:quantitative-results}). A case study is further detailed in Appendix~\ref{sec:case1}.

\subsection{Evaluation Setup}
\label{subsec:evaluation-setup}

\paragraph{Datasets.}
\begin{figure}[t]
    \centering
    \includegraphics[width=1\linewidth]{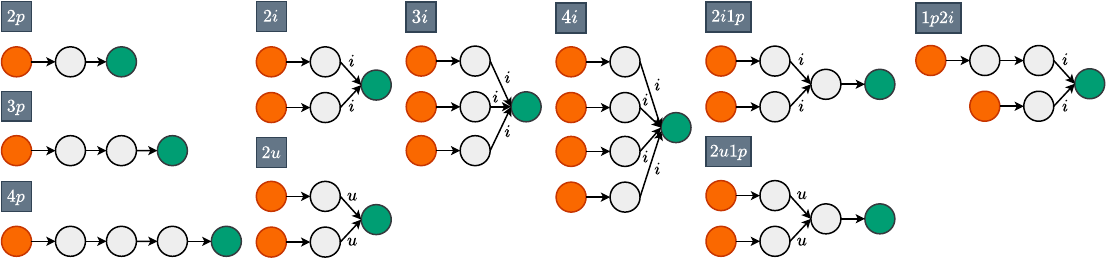}
    \caption{Illustration of the query structures considered in our experiments, following the representation of Ren et al.~\cite{Ren2024Neural}. Orange nodes (left) denote anchor entities, middle nodes denote existential variables, and green nodes (right) denote answer variables. In the query names, $p$, $i$, and $u$ represent projection, intersection, and union, respectively. }
    \label{fig:querytypes}
\end{figure}

We conduct experiments on FB15k-237~\cite{Toutanova2015Observed} and NELL995~\cite{Xiong2017DeepPath} using the same splits as Ren et al.~\cite{Ren2020Query2box}, and additionally on the harder variants FB15k-237+H and NELL995+H~\cite{Gregucci2025Complex}. Fig.~\ref{fig:querytypes} illustrates all query types included in these datasets ($4p$ and $4i$ queries exist only in the +H datasets). Further details and statistics are provided in Appendix~\ref{sec:appendixdataset}.

\paragraph{Complex Query Answering Approaches.}
We use the CQD neurosymbolic approach to execute complex queries, as it forms the basis of other methods such as  $\mathrm{CQD}^{\mathcal{A}}$~\cite{Arakelyan2023Adapting} and QTO~\cite{Bai2023Answering}. Therefore, CQD-SHAP can also explain the outputs of these models in a similar manner, as they differ mainly in score normalization or search pruning strategies. Implementation details and hyperparameter settings are provided in the Appendix~\ref{sec:cqa-model}.

\paragraph{Performance Metrics.}
\label{subsec:performance-metrics}

Following CQD, the atom containing the target variable $V_A$ always produces a score for every entity in the graph. We adopt Mean Reciprocal Rank (MRR) as the evaluation metric, following its standard definition~\cite{Rossi2021Knowledge}, to assess how well the model predicts hard answers $e_i \in \mathcal{E}_{\mathrm{hard}}$. The formal definition of MRR is provided in Appendix~\ref{sec:metrics}. The metric is computed in a filtered setting~\cite{Bordes2013Translating}, as formally defined in Equation~\eqref{eq:filteredlist}.

\paragraph{Evaluation Scenarios.}
\label{subsec:explanation-metrics}

To evaluate the effectiveness of our explanation method, we extend the notions of \emph{necessary} and \emph{sufficient} explanations~\cite{Rossi2022Explaining} from link prediction to the context of explaining atoms in complex queries. In this extension, we evaluate all query–answer pairs, whereas Rossi et al.~\cite{Rossi2022Explaining} restrict their evaluation to pairs for which the original model ranks the target entity at the top position in the necessary setting and fails to do so in the sufficient setting, making our evaluation more rigorous.

\emph{Necessary Explanations:}
Given a query $\mathcal{Q}_S$ and a hard answer $e_i \in \mathcal{E}_{\mathrm{hard}}$, we extract explanations in terms of the query’s atoms. An explanation is considered necessary if, upon executing the most important atom in the explanation symbolically, the hard answer $e_i$ drops in the ranking. The effectiveness of a necessary explanation is thus measured by the decrease in MRR when this atom is replaced by symbolic reasoning. We repeat this process for each hard answer in a single query, computing the change in performance metrics for each, and then average these values to obtain the metrics for the query. Finally, we report the average effectiveness across all queries of a given type.

\emph{Sufficient Explanations:}
Complementarily, an explanation is sufficient if executing only the most important atom identified by the explanation using the neural approach increases MRR compared to a fully symbolic baseline. Averaging is performed over hard answers and queries as in the necessity setting.

Formal definitions of these evaluation scenarios are provided in Appendix~\ref{sec:evaluationscenarios}.

\paragraph{Explanation Methods.}
\label{subsec:baselines}

As there are no existing methods to assign importance to atoms in complex queries, we introduce several baselines to enable comparison with our approach in terms of necessary and sufficient explanations: 
(1)~\emph{First:} The first level of the query execution graph refers to the atoms that contain an anchor entity (e.g., $a_1$ in Fig.~\ref{fig:2p}). In this baseline, we randomly select one atom from the first level and execute it differently from the others—symbolically in the necessary scenario and neurally in the sufficient scenario.
(2)~\emph{Last:} This baseline is defined similarly to the first one, except that the randomly selected atom is chosen from the last level of the query graph (i.e., atoms containing the target variable).
(3)~\emph{Random:} One atom is selected randomly and executed differently from the other atoms.
(4)~\emph{Score-based:} The query is first run using the neurosymbolic approach, and the link prediction scores assigned to each atom along a path to the target answer are collected. For most query types, the atom with the lowest link prediction score is selected for different execution, based on the assumption that lower scores indicate greater importance of missing link prediction, while higher scores typically reflect observed links. However, when there is a union operation between atoms, such as in the $2u$ and $2u1p$ queries, we select the higher score between the atoms involved in the union, reflecting the use of t-conorms. 
(5)~\emph{CQD-SHAP:} Our proposed method computes Shapley values for all atoms with respect to a target answer and selects the atom with the highest Shapley value for different execution.

\begin{table*}[!t]
\centering
\caption{
Explanation necessity and sufficiency on each dataset for different query types and baselines, measured by changes in mean reciprocal rank ($\Delta$MRR, in \%). Lower is better for necessity, higher is better for sufficiency.}
\label{tab:combined_eval}
\scriptsize
\begin{tabular}{lrrrrrrrrrr}
\toprule
\raisebox{-0.8ex}{\textbf{Query}}
& \multicolumn{5}{c}{\textbf{Necessary}}
& \multicolumn{5}{c}{\textbf{Sufficient}} \\
\cmidrule(lr){2-6} \cmidrule(lr){7-11}
& First & Last & Random & Score & \textbf{CQD-SHAP}
& First & Last & Random & Score & \textbf{CQD-SHAP} \\
\midrule
\multicolumn{11}{c}{\textbf{FB15k-237}} \\
\midrule
{\boldmath$2p$}    & -2.60 & -5.41 & -4.34 & -7.62 & \textbf{-15.24} & +19.17 & +21.98 & +20.90 & +24.18 & \textbf{+31.81} \\
{\boldmath$3p$}    & -1.00 & -3.01 & -0.84 & -4.71 & \textbf{-11.14} & +11.62 & +16.64 & +13.13 & +18.13 & \textbf{+26.68} \\
{\boldmath$2i$}    & -4.11 & -4.11 & -4.11 & -18.24 & \textbf{-26.27} & +18.69 & +18.69 & +18.69 & +32.82 & \textbf{+40.86} \\
{\boldmath$3i$}    & -5.55 & -5.55 & -5.55 & -22.42 & \textbf{-37.14} & +10.87 & +10.87 & +10.87 & +23.49 & \textbf{+42.27} \\
{\boldmath$2u$}    & -7.20 & -7.20 & -7.20 & -24.25 & \textbf{-24.32} & +17.72 & +17.72 & +17.72 & +34.78 & \textbf{+34.84} \\
{\boldmath$2u1p$}  & +1.13 & -8.81 & -2.21 & +2.33 & \textbf{-12.89} & +5.19 & +20.71 & +10.17 & +10.64 & \textbf{+24.38} \\
{\boldmath$1p2i$}  & +0.01 & -1.46 & -0.89 & -8.70 & \textbf{-16.29} & +8.66 & +12.98 & +10.53 & +21.23 & \textbf{+29.27} \\
{\boldmath$2i1p$}  & -1.00 & -5.16 & -2.06 & -6.50 & \textbf{-12.67} & +7.61 & +13.92 & +9.53 & +15.11 & \textbf{+22.32} \\
\midrule
\multicolumn{11}{c}{\textbf{FB15k-237+H}} \\
\midrule
{\boldmath$2p$}    & -1.03 & -2.69 & -1.57 & -3.01 & \textbf{-3.87} & +1.51 & +3.18 & +2.06 &  +3.50  & \textbf{+4.35} \\
{\boldmath$3p$}    & -0.59 & -1.73 & -0.83 & -1.74 & \textbf{-2.24} & +0.53 & +1.68 & +1.15 &  +1.53  & \textbf{+2.43} \\
{\boldmath$4p$}    & -1.49 & -1.78 & -1.53 & -2.11 & \textbf{-2.56} & +0.42 & +1.21 & +0.85 &  +1.29  & \textbf{+2.31} \\
{\boldmath$2i$}    & -7.20 & -7.20 & -7.20 & -7.05 & \textbf{-8.29} & +4.61 & +4.61 & +4.61 &  +4.46  & \textbf{+5.70} \\
{\boldmath$3i$}    & -6.08 & -6.08 & -6.08 & -6.75 & \textbf{-9.07} & +4.09 & +4.09 & +4.09 &  +3.85  & \textbf{+6.04} \\
{\boldmath$4i$}    & -7.49 & -7.49 & -7.49 & -11.49 & \textbf{-16.61} & +4.34 & +4.34 & +4.34 & +4.05  & \textbf{+7.99} \\
{\boldmath$2u$}    & -10.98 & -10.98 & -10.98 & -19.54 & \textbf{-23.44} & +31.65 & +31.65 & +31.65 & +40.21 & \textbf{+44.12} \\
{\boldmath$2u1p$}  & -1.61 & -0.84 & -1.51 & -1.25 & \textbf{-4.25} & +3.67 & +4.92 & +4.24 &  +4.08  & \textbf{+8.31} \\
{\boldmath$1p2i$}  & -1.28 & -0.98 & -1.11 & -1.03 & \textbf{-3.65} & +1.85 & +2.12 & +1.91 &  +1.68  & \textbf{+5.02} \\
{\boldmath$2i1p$}  & -1.31 & -3.60 & -1.83 & -2.99 & \textbf{-7.89} & +3.92 & +5.47 & +4.47 &  +4.83  & \textbf{+10.68} \\
\midrule
\multicolumn{11}{c}{\textbf{NELL995}} \\
\midrule
{\boldmath$2p$}    &  -6.80 & -9.22 & -7.83 & -16.40 & \textbf{-23.33} & +22.25 & +24.67 & +23.27 & +31.84 & \textbf{+38.77} \\
{\boldmath$3p$}    & -5.19 & -3.13 & -2.82 & -10.94 & \textbf{-16.31} & +17.26 & +18.37 & +16.78 & +27.60 & \textbf{+35.47} \\
{\boldmath$2i$}    & -5.75 & -5.75 & -5.75 & -19.39 & \textbf{-26.92} & +14.75 & +14.75 & +14.75 & +28.38 & \textbf{+35.91} \\
{\boldmath$3i$}    & -7.17 & -7.17 & -7.17 & -25.92 & \textbf{-39.64} & +9.27 & +9.27 & +9.27 & +23.07 & \textbf{+38.25} \\
{\boldmath$2u$}    & -20.30 & -20.30 & -20.30 & -45.67 & \textbf{-45.70} & +26.76 & +26.76 & +26.76 & +52.11 & \textbf{+52.13} \\
{\boldmath$2u1p$}  & +3.21 & -3.78 & +0.61 & +7.29 & \textbf{-9.18} & +7.45 & +25.30 & +13.15 & +14.88 & \textbf{+30.71} \\
{\boldmath$1p2i$}  & -0.05 & -1.55 & -1.31 & -9.85 & \textbf{-17.65} & +8.75 & +11.99 & +10.10 & +18.90 &  \textbf{+25.27}\\
{\boldmath$2i1p$}  &  -1.17 & -8.02 & -3.27 & -12.22 & \textbf{-17.27} & +6.28 & +15.54 & +9.12 & +18.04 & \textbf{+24.01} \\
\midrule
\multicolumn{11}{c}{\textbf{NELL995+H}} \\
\midrule
{\boldmath$2p$}    & -4.60 & -6.08 & -5.48 & -7.34 & \textbf{-8.52} & +2.90 & +4.38 & +3.78 & +5.64 & \textbf{+6.82} \\
{\boldmath$3p$}    & -2.88 & -2.81 & -2.56 & -4.39 & \textbf{-4.68} & +1.71 & +2.02 & +1.72 & +2.82 & \textbf{+4.19} \\
{\boldmath$4p$}    & -3.03 & -3.00 & -2.93 & -3.83 & \textbf{-4.49} & +0.79 & +1.43 & +0.99 & +1.98 & \textbf{+3.20} \\
{\boldmath$2i$}    & -12.11 & -12.11 & -12.11 & -11.45 & \textbf{-14.52} & +7.88 & +7.88 & +7.88 & +7.23 & \textbf{+10.29} \\
{\boldmath$3i$}    & -8.84 & -8.84 & -8.84 & -10.48 & \textbf{-15.09} & +4.84 & +4.84 & +4.84 & +4.74 & \textbf{+8.33} \\
{\boldmath$4i$}    & -7.42 & -7.42 & -7.42 & -10.60 & \textbf{-18.22} & +3.90 & +3.90 & +3.90 & +4.07 & \textbf{+8.84} \\
{\boldmath$2u$}    & -12.56 & -12.56 & -12.56 & -23.41 & \textbf{-27.29} & +44.16 & +44.16 & +44.16 & +55.02 & \textbf{+58.90} \\
{\boldmath$2u1p$}  & -3.34 & -4.16 & -3.68 & -2.51 & \textbf{-8.15} & +4.62 & +8.76 & +6.37 & +5.45 & \textbf{+13.12} \\
{\boldmath$1p2i$}  & +0.34 & +1.02 & -0.78 & -2.17 & \textbf{-9.48} & +4.98 & +6.51 & +6.67 & +8.07 & \textbf{+14.12} \\
{\boldmath$2i1p$}  & -3.55 & -12.56 & -6.51 & -10.28 & \textbf{-15.94} & +3.22 & +11.72 & +5.91 & +9.43 & \textbf{+15.16} \\
\bottomrule
\end{tabular}
\end{table*}

\subsection{Quantitative Results}
\label{subsec:quantitative-results}

Table~\ref{tab:combined_eval} shows the quantitative evaluation results for different query types and datasets under each evaluation scenario
as produced by the explanation methods defined in Section~\ref{subsec:baselines}. Since $2u$, $2i$, $3i$ and $4i$ queries have only one level of atoms, the \emph{First} and \emph{Last} results are the same as \emph{Random} for these types. CQD-SHAP outperforms the baselines in all cases. Considering all datasets, the explanations produced by CQD-SHAP reduce MRR by between 2.24\% (for $3p$ on FB15k-237+H) and 45.70\% (for $2u$ on NELL995) in the necessary scenario, and increase MRR by between 2.31\% (for $4p$ on FB15k-237+H) and 58.90\% (for $2u$ on NELL995+H) in the sufficient scenario, depending on the query type.

Comparing CQD-SHAP with the best baseline in each case, CQD‑SHAP demonstrates a clearly stronger effect. Most notably, for $3i$ queries on FB15k‑237, the best baseline (\emph{Score}) achieves -22.42\% in the necessary and +23.49\% in the sufficient evaluation, while CQD‑SHAP reaches -37.14\% and +42.27\%, an additional decrease of 14.72 percentage points in the necessary setting and an additional increase of 18.78 percentage points in the sufficient setting. On NELL995, the additional decrease and increase over the best baseline (\emph{Score}) are 13.72 and 15.18 percentage points, respectively.

When comparing standard and harder benchmarks, the absolute $\Delta$MRR values are generally smaller on the +H variants, which is expected as the harder benchmarks reduce the proportion of queries reducible to simpler problems. For example, in the standard benchmark many $3p$ queries have only one missing link; therefore, identifying this link through the explanation method and executing it symbolically leads to a more significant drop in ranking. In the harder benchmarks, by contrast, only a subset of queries can be reduced this way, while others require two or all links to be inferred. Consequently, modifying only one atom has a smaller effect. A similar behavior is observed in the sufficiency setting, where executing only one atom neurally leads to a smaller improvement in ranking when queries are less reducible and require more complex reasoning.

For CQD-SHAP, explanation effectiveness is consistently greater for intersection queries ($2i$, $3i$, $4i$) than for projection queries ($2p$, $3p$, $4p$), in terms of both larger reductions in the necessary evaluation and larger improvements in the sufficient evaluation. For example, on FB15k-237 the necessary and sufficient metrics for the $3p$ query type are -11.14\% and +26.68\%, versus -37.14\% and +42.27\% for $3i$. A similar difference holds on the harder variant, from -2.24\% and +2.43\% for $3p$ to -9.07\% and +6.04\% for $3i$, and likewise for NELL995 and its harder version. Upon further investigation, we find that this behavior arises from structural differences between these query types. In projection queries, multiple paths may lead to the correct answer, so executing one atom differently still allows others to recover it via alternative paths. In contrast, in intersection queries, answers are aggregated via t-norm (e.g., multiplication of scores) across all branches. As a result, the most important branch acts as a critical filter, and executing it differently has a substantially stronger impact on the final ranking. Overall, this indicates that CQD-SHAP captures these structural characteristics of query types, whereas the other baselines do not consistently reflect this pattern.

\section{Discussion}
\label{sec:discussion}

Examining the baseline results more closely, the baselines not only fail to consistently outperform each other, but in some cases produce positive scores in the necessary evaluation, meaning the ranking improves when executing the selected atom symbolically. For example, for each baseline, we observe +3.21, +0.61, and +7.29 for the necessary evaluation of the \emph{First}, \emph{Random}, and \emph{Score} baselines on the NELL995 dataset for the $2u1p$ query type, and +1.02 for the \emph{Last} baseline on the NELL995+H dataset for the $1p2i$ query type. However, CQD-SHAP always yields negative values in the necessary evaluation, with a minimum drop of -2.24.

Although computing Shapley values is NP-hard, in our case, the number of players (query atoms) is small (at most four), which keeps the computation efficient even when computing exact Shapley values without approximation. In our experiments, the average computation time per query–answer pair was 30.92~ms on FB15k-237 and 102.67~ms on NELL995, rising to 57.46~ms and 140.38~ms for the +H variants due to the additional $4i$ and $4p$ queries. Detailed runtime results per query type and dataset are reported in Appendix~\ref{sec:runtime}.

An interesting finding is that the most important atom identified by CQD-SHAP is not always the one corresponding to the missing link. It can instead be an atom whose triple already exists in the KG. Consider a $2p$ query asking \textit{``Which films were distributed by companies that are subsidiaries of Time Warner?''} with target hard answer \textit{The Animatrix}. The KG already contains the fact that this film was distributed by \textit{Warner Home Video}, but the \textit{child} relation between \textit{Time Warner} and \textit{Warner Home Video} is missing. One might expect CQD-SHAP to flag the first atom as most important, yet it assigns importance scores of $-166.0$ and $+9{,}963.0$ to the first and second atoms respectively. The negative score for the first atom suggests that neural execution of this step may hurt the ranking of \textit{The Animatrix} in some cases, because the neural model may introduce noise (e.g., produce incorrect subsidiary candidates, elevating false answers above \textit{Warner Home Video}). The second atom receives a vastly higher score for a more subtle reason. The KG contains 14 other subsidiaries of \textit{Time Warner}, such as \textit{Warner Bros. Entertainment}, which belong to the same corporate family as \textit{Warner Home Video} and are thus likely nearby in the latent space. This embedding proximity allows the neural model to assign a high score to \textit{The Animatrix} via these alternative subsidiaries, even without predicting the exact missing link. The second atom thus acts as the primary driver of its ranking, which CQD-SHAP faithfully captures. Further details are provided in Appendix~\ref{sec:case2}.

Future work could focus on redefining the goal of explanations in CQA. For example, instead of working at the query atom level, it could aim to identify the most influential triples in the KG that contribute to a target answer. Even when keeping the focus on query atoms, alternative definitions of value functions in the Shapley game could be explored. Furthermore, future work could focus on developing global explanations, for example, by aggregating the local explanations produced by our work to obtain a broader understanding of the model's behavior. Finally, integrating an LLM- or template-based natural language generation module could convert Shapley values into human-readable explanations, e.g., stating that \texttt{Insulin} ranks highly mainly due to the strong contribution of the atom ``causes kidney toxicity'', while ``prescribed for diabetes'' has a weaker effect.

\section{Conclusion}

We defined an explanation goal in the context of complex query answering and presented CQD‑SHAP, a Shapley‑based method that quantifies each query atom's contribution to the ranking of a target answer by assessing the benefit of neural execution over symbolic retrieval. We also redefined the concepts of necessary and sufficient explanations for this domain and demonstrated CQD‑SHAP's effectiveness through quantitative evaluations on two well‑studied knowledge graphs and two query generation methods per KG (original version and +H variant) achieving the best performance across all datasets and query types. We further investigated the query structures and discussed them in more detail to show how CQD‑SHAP captures these characteristics. We also discussed with an example that the most important atom is not always a missing link. Finally, we suggested several directions for future work, including redefining explanation goals, designing alternative Shapley games, developing global explanations beyond single‑answer interpretations, and integrating natural language generation to interpret Shapley values at a human‑readable level.

\section*{Acknowledgments}
The authors used LLMs (GPT-5.2 by OpenAI and Claude Sonnet 4.6 by Anthropic) to assist with grammar correction and improving the readability of the text. The authors take full responsibility for all content of this paper.

\bibliographystyle{splncs04}
\bibliography{bibliography}

\appendix
\section{Datasets}
\label{sec:appendixdataset}

This section describes the knowledge graphs and query benchmarks used in our evaluation, including an overview of the datasets (Section~\ref{sec:dataset-overview}), knowledge graph statistics (Section~\ref{sec:kg-stats}), query structures (Section~\ref{sec:query-structures}), and query statistics (Section~\ref{sec:query-stats}).

\subsection{Benchmark Datasets}
\label{sec:dataset-overview}

Similar to CQD~\cite{Arakelyan2021CQD}, we use the FB15k-237~\cite{Toutanova2015Observed} and NELL995~\cite{Xiong2017DeepPath} knowledge graphs, with queries generated by Ren et al.~\cite{Ren2020Query2box}. Although widely used for CQA, these benchmarks exhibit limited compositional difficulty: for several query types, most queries (up to 98\%) can be reduced to simpler inference tasks such as single-link prediction~\cite{Gregucci2025Complex}, meaning that performing neural inference for only one link is sufficient to obtain the desired answer.

Gregucci et al.~\cite{Gregucci2025Complex} computed these statistics by analyzing whether an answer remains reachable when training triples are used for some atoms and test triples for others. For example, for the $2p$ query type, if the answer is reachable when training triples are used for one atom and test triples for the other, the query is considered reducible to $1p$. If the answer is reachable only when test triples are used for both atoms, the query is considered non-reducible and remains a true $2p$ query. They found that most query types are not balanced in terms of reducibility; for instance, 98.1\% of the query-answer pairs in $2p$ are reducible to $1p$, whereas only 1.9\% are genuinely $2p$.

To address this issue, they generated additional query-answer pairs to balance reducibility levels and introduced the FB15k-237+H and NELL995+H datasets. In these datasets, for example, the $2p$ query type contains exactly 10,000 query–answer pairs reducible to $1p$ and 10,000 non-reducible pairs that remain true $2p$ queries. To increase complexity, they also introduced the $4i$ and $4p$ query types. We evaluate our explanation method on these harder benchmarks to assess its effectiveness in a more balanced and challenging setting.

It should also be noted that the new benchmarks include negation query types. However, since CQD does not natively support negation and its implementation requires score calibration, we exclude these types from our evaluation.

\begin{table}[h]
\caption{Statistics of the knowledge graphs used in our experiments, including the number of nodes, relations, and edges in the training, validation, and test graphs.}
\label{tab:kg-stats}
\scriptsize
\setlength{\tabcolsep}{6pt}
\centering
\begin{tabular}{@{}lrrrrr@{}}
\toprule
\textbf{Dataset} & \textbf{Nodes} & \textbf{Relations} & $\mathbf{|G_{\mathrm{train}}|}$ & $\mathbf{|G_{\mathrm{valid}}|}$ & $\mathbf{|G_{\mathrm{test}}|}$ \\
\midrule
\textbf{FB15k-237} & 14{,}505 & 474 & 544{,}230 & 579{,}300 & 620,232 \\
\textbf{NELL995}   & 63{,}361 & 400 & 456{,}852 & 716{,}472 & 1{,}005{,}086 \\
\bottomrule
\end{tabular}
\end{table}

\subsection{Knowledge Graph Statistics}
\label{sec:kg-stats}

Both the original and the harder (+H) variants use the same knowledge graph, and the only difference lies in the validation and test queries. The numbers of nodes, relations, and edges in the training graph ($G_{\mathrm{train}}$), validation graph ($G_{\mathrm{valid}}$), and test graph ($G_{\mathrm{test}}$) are reported in Table~\ref{tab:kg-stats}.

\begin{table}[!htbp]
\caption{Formal logical definitions of all query types used in our evaluation.}
\label{tab:query-structures}
\scriptsize
\setlength{\tabcolsep}{11.5pt}
\centering
\begin{tabular}{@{}l l@{}}
\toprule
\textbf{Type} & \textbf{Logic} \\
\midrule
$2p$   & $?V_2: \exists V_1 \cdot p_1(e_1, V_1) \wedge p_2(V_1, V_2)$ \\
$3p$   & $?V_3: \exists V_1, V_2 \cdot p_1(e_1, V_1) \wedge p_2(V_1, V_2) \wedge p_3(V_2, V_3)$ \\
$4p$   & $?V_4: \exists V_1, V_2, V_3 \cdot p_1(e_1, V_1) \wedge p_2(V_1, V_2) \wedge p_3(V_2, V_3) \wedge p_4(V_3, V_4)$ \\
$2i$   & $?V_1: p_1(e_1, V_1) \wedge p_2(e_2, V_1)$ \\
$3i$   & $?V_1: p_1(e_1, V_1) \wedge p_2(e_2, V_1) \wedge p_3(e_3, V_1)$ \\
$4i$   & $?V_1: p_1(e_1, V_1) \wedge p_2(e_2, V_1) \wedge p_3(e_3, V_1) \wedge p_4(e_4, V_1)$\\
$2u$   & $?V_1: p_1(e_1, V_1) \vee p_2(e_2, V_1)$ \\
$2u1p$ & $?V_2: \exists V_1 \cdot (p_1(e_1, V_1) \vee p_2(e_2, V_1)) \wedge p_3(V_1, V_2)$ \\
$1p2i$ & $?V_2: \exists V_1 \cdot p_1(e_1, V_1) \wedge p_2(V_1, V_2) \wedge p_3(e_2, V_2)$ \\
$2i1p$ & $?V_2: \exists V_1 \cdot p_1(e_1, V_1) \wedge p_2(e_2, V_1) \wedge p_3(V_1, V_2)$ \\
\bottomrule
\end{tabular}
\end{table}

\subsection{Query Structures}
\label{sec:query-structures}

The complex queries used in our experiments are built upon three standard operations: relation projection ($p$), intersection ($i$), and union ($u$). The type of each query specifies which operations are involved, starting from the anchor entities, as well as the number of projection hops or branches to be merged. For example, the $2i1p$ query type indicates that two branches are first combined via intersection, followed by a projection over the resulting set. Note that each branch in a union or intersection operation is itself a projection. For instance, in a $2i$ query, there is a projection from each anchor entity (two in total), followed by their intersection.

Query types involving only projection and intersection ($2p$, $3p$, $2i$, $3i$, $2i1p$, $1p2i$) were first introduced by Hamilton et al.~\cite{Hamilton2018Embedding}, while query types involving unions ($2u$, $2u1p$) were first introduced by Ren et al.~\cite{Ren2020Query2box}. Recently, Gregucci et al.~\cite{Gregucci2025Complex} introduced two more complex query types, $4p$ and $4i$, in their harder benchmark (+H variants), which were not present in the original benchmark.

While each query type has been visually presented in Fig.~\ref{fig:querytypes} in Section~\ref{subsec:evaluation-setup}, we provide their formal logical definitions in Table~\ref{tab:query-structures}. In each formula, free variables (prefixed with $?$) denote the target variables whose groundings constitute the answer set, while existentially quantified variables (bounded by $\exists$) must be bound to entities along the reasoning path. Anchor entities, i.e., the known starting points of a query, are denoted by $e_i$ and each $p_i(e_j, V_k)$ or $p_i(V_j, V_k)$ is called an atom of the query. Each atom is a projection of predicate $p_i$ from the subject of that atom ($e_j$, or bounded $V_j$). Intersection and union queries share the same variable in the object of all branch atoms (e.g., $V_1$ in $2i$ or $2u$ query types). However, in chain projection queries (e.g., $2p$, $3p$, and $4p$) there is a sequence of variables where the object variable of each atom is the subject variable of the next one (e.g., $V_1$ is the object of the first atom and the subject of the second atom in the $2p$ query).

\begin{table}[!htbp]
\centering
\caption{Query statistics per query type and dataset. \textit{Queries} denotes the number of queries evaluated, \textit{Pairs} the total number of query-answer pairs, and \textit{Avg} the average number of hard answer per query.}
\label{tab:query_stats}
\setlength{\tabcolsep}{2.8pt}
\fontsize{6pt}{7pt}\selectfont
\begin{tabular}{@{}l@{\hskip -5pt}l rrr rrr rrr rrr@{}}
\toprule
\textbf{Type}& & \multicolumn{3}{c}{\textbf{FB15k-237}} & \multicolumn{3}{c}{\textbf{FB15k-237+H}} & \multicolumn{3}{c}{\textbf{NELL995}} & \multicolumn{3}{c}{\textbf{NELL995+H}} \\
\cmidrule(lr){3-5} \cmidrule(lr){6-8} \cmidrule(lr){9-11} \cmidrule(lr){12-14}
 & & \textbf{Queries} & \textbf{Pairs} & \textbf{Avg} & \textbf{Queries} & \textbf{Pairs} & \textbf{Avg} & \textbf{Queries} & \textbf{Pairs} & \textbf{Avg} & \textbf{Queries} & \textbf{Pairs} & \textbf{Avg} \\
\midrule
$2p$   & & 5,000 & 98,551  & 19.71 & 936   & 20,000 & 21.36 & 4,000 & 36,357  & 9.08  & 1,054  & 20,000 & 18.97 \\
$2u$   & & 5,000 & 17,294  & 3.45  & 6,695  & 10,000 & 1.49  & 4,000 & 8,311   & 2.07  & 8,168  & 10,000 & 1.22  \\
$2i$   & & 5,000 & 48,917  & 9.78  & 6,692  & 20,000 & 2.98  & 4,000 & 29,880  & 7.47  & 8,278  & 20,000 & 2.41  \\
$3i$   & & 5,000 & 37,745  & 7.54  & 9,615  & 30,000 & 3.12  & 4,000 & 23,054  & 5.76  & 13,118 & 30,000 & 2.28  \\
$3p$   & & 5,000 & 159,699 & 31.93 & 1,461  & 30,000 & 20.53 & 4,000 & 62,323  & 15.58 & 1,583  & 30,000 & 18.95 \\
$2u1p$ & & 5,000 & 56,430  & 11.28 & 473   & 30,000 & 63.42 & 4,000 & 34,597  & 8.64  & 1,073  & 30,000 & 27.95 \\
$2i1p$ & & 5,000 & 202,531 & 40.50 & 1,586  & 40,000 & 25.22 & 4,000 & 133,736 & 33.43 & 1,957  & 40,000 & 20.43 \\
$1p2i$ & & 5,000 & 347,713 & 69.54 & 604   & 40,000 & 66.22 & 4,000 & 209,554 & 52.38 & 1,260  & 40,000 & 31.74 \\
$4i$   & & ---  & ---    & ---   & 20,189 & 40,000 & 1.98  & ---  & ---    & ---   & 22,326 & 40,000 & 1.79  \\
$4p$   & & ---  & ---    & ---   & 2,670  & 40,000 & 14.98 & ---  & ---    & ---   & 2,322  & 40,000 & 17.22 \\
\bottomrule
\end{tabular}%
\end{table}

\subsection{Query Statistics}
\label{sec:query-stats}

Table~\ref{tab:query_stats} provides an overview of the query statistics across all four datasets and ten query types used in our evaluation.
The original benchmark (FB15k-237 and NELL995) evaluates eight query types with a fixed number of 5{,}000 and 4{,}000 queries respectively, while the harder benchmark (FB15k-237+H and NELL995+H) extends this with two additional query types ($4i$ and $4p$) and instead fixes the total number of answer pairs per query type, resulting in a variable number of queries.
The average number of answer pairs per query varies considerably across query types: intersection queries ($2i$, $3i$, $4i$) tend to have fewer answer pairs per query due to the conjunctive constraints.

In most cases, except for the $2u$ and $3i$ query types in NELL-based datasets, the original benchmark has a higher number of query-answer pairs than the harder variants, most notably for the $1p2i$ query type, where there are 347{,}713 pairs in FB15k-237 compared to 40{,}000 pairs in FB15k-237+H.
However, as mentioned in Section~\ref{sec:appendixdataset}, most of the pairs in the original dataset are reducible to easier types.
For example, 86.8\% of the $1p2i$ queries in FB15k-237 are reducible to just $1p$ queries (i.e., there is only one missing link in them), whereas in the harder datasets this is balanced, with only 25\% of such cases in FB15k-237+H for this query type.

\section{Implementation Details}
\label{sec:impdetails}

We describe the implementation of our method, including the complex query answering model (Section~\ref{sec:cqa-model}), the performance metric used for evaluation (Section~\ref{sec:metrics}), and the hardware setup (Section~\ref{sec:hardware}).

\subsection{Complex Query Answering Model}
\label{sec:cqa-model}

We use the CQD method to retrieve the answers to complex queries. Since our methodology requires executing some parts of a query neurally while handling others symbolically, we reimplement the combinatorial query answering procedure proposed in CQD. We validate our implementation by obtaining results consistent with the original CQD across all query types.
For neural atom execution, we use the pre-trained link prediction model provided by the CQD authors without additional fine-tuning. We adopt a default beam width of $k=10$ for all query types, together with the product t-norm $\top_{\text{prod}}(s_1, s_2)=s_1 \cdot s_2$ and the product t-conorm $\perp_{\text{prod}}(s_1, s_2)=(s_1 + s_2)-(s_1 \cdot s_2)$. All code and required resources (e.g., datasets, pre-trained models) are provided in our GitHub repository.\footnote{\url{https://github.com/ds-jrg/CQD-SHAP}}

\subsection{Performance Metrics}
\label{sec:metrics}

\paragraph{Mean Reciprocal Rank (MRR):}
Given a single query $\mathcal{Q}_S$ with multiple hard answers $\mathcal{E}_{\mathrm{hard}}$, MRR is the average inverse rank of each hard answer $e_i \in \mathcal{E}_{\mathrm{hard}}$:
\begin{align}
\text{MRR} = \frac{1}{|\mathcal{E}_{\mathrm{hard}}|} \sum_{e_i \in \mathcal{E}_{\mathrm{hard}}} \frac{1}{r_i(\mathcal{Q}_S)}
\label{eq:metricmrr}
\end{align}
where $r_i(\mathcal{Q}_S)$ denotes the rank of $e_i$ in the filtered answer set (Equation~\eqref{eq:rankeq}).

\subsection{Evaluation Scenarios}
\label{sec:evaluationscenarios}
We here provide the formal definitions of the necessary and sufficient evaluation scenarios introduced in Section~\ref{subsec:evaluation-setup}. Let $\mathcal{Q}_t$ denote the set of all queries of type $t$, $\mathcal{A} = \{a_1, \ldots, a_{|\mathcal{A}|}\}$ the set of all atoms in a query $\mathcal{Q}$, $\mathcal{E}_{\mathrm{hard}}^{\mathcal{Q}}$ the set of hard answers for query $\mathcal{Q}$, and $a^*_i$ the most important atom with respect to hard answer $e_i$:
\begin{align}
a^*_i = \arg\max_{a \in \mathcal{A}} \phi_a(e_i)
\label{eq:most-important-atom}
\end{align}
For the \emph{necessary evaluation}, we measure the MRR when all atoms are executed neurally except $a^*_i$, which is instead executed symbolically ($\text{MRR}_{\text{nec}}$), and compare it against the fully neural baseline ($\text{MRR}_{\text{neur}}$). The difference between the two, $\Delta\text{MRR}_{\text{nec}}$ (Equation~\eqref{eq:delta-mrr-nec}), reflects how necessary $a^*_i$ is to the model's performance, where a negative value confirms its necessity.
\begin{align}
\text{MRR}_{\text{neur}} &= \frac{1}{|\mathcal{Q}_t|} \sum_{\mathcal{Q} \in \mathcal{Q}_t} \frac{1}{|\mathcal{E}_{\mathrm{hard}}^{\mathcal{Q}}|} \sum_{e_i \in \mathcal{E}_{\mathrm{hard}}^{\mathcal{Q}}} \frac{1}{r_i(\mathcal{Q}_{\mathcal{A}})}
\label{eq:mrr-neur}
\end{align}
\begin{align}
\text{MRR}_{\text{nec}} &= \frac{1}{|\mathcal{Q}_t|} \sum_{\mathcal{Q} \in \mathcal{Q}_t} \frac{1}{|\mathcal{E}_{\mathrm{hard}}^{\mathcal{Q}}|} \sum_{e_i \in \mathcal{E}_{\mathrm{hard}}^{\mathcal{Q}}} \frac{1}{r_i(\mathcal{Q}_{\mathcal{A} \setminus \{a^*_i\}})}
\label{eq:mrr-nec}
\end{align}
\begin{align}
\Delta\text{MRR}_{\text{nec}} &= \text{MRR}_{\text{nec}} - \text{MRR}_{\text{neur}}
\label{eq:delta-mrr-nec}
\end{align}
For the \emph{sufficient evaluation}, we measure the MRR when only $a^*_i$ is executed neurally while all remaining atoms are executed symbolically ($\text{MRR}_{\text{suf}}$), and compare it against the fully symbolic baseline ($\text{MRR}_{\text{sym}}$). The difference between the two, $\Delta\text{MRR}_{\text{suf}}$ (Equation~\eqref{eq:delta-mrr-suf}), reflects how sufficient $a^*_i$ alone is to improve the model's performance, where a positive value confirms its sufficiency.
\begin{align}
\text{MRR}_{\text{sym}} &= \frac{1}{|\mathcal{Q}_t|} \sum_{\mathcal{Q} \in \mathcal{Q}_t} \frac{1}{|\mathcal{E}_{\mathrm{hard}}^{\mathcal{Q}}|} \sum_{e_i \in \mathcal{E}_{\mathrm{hard}}^{\mathcal{Q}}} \frac{1}{r_i(\mathcal{Q}_{\{\}})}
\label{eq:mrr-sym}
\end{align}
\begin{align}
\text{MRR}_{\text{suf}} &= \frac{1}{|\mathcal{Q}_t|} \sum_{\mathcal{Q} \in \mathcal{Q}_t} \frac{1}{|\mathcal{E}_{\mathrm{hard}}^{\mathcal{Q}}|} \sum_{e_i \in \mathcal{E}_{\mathrm{hard}}^{\mathcal{Q}}} \frac{1}{r_i(\mathcal{Q}_{\{a^*_i\}})}
\label{eq:mrr-suf}
\end{align}
\begin{align}
\Delta\text{MRR}_{\text{suf}} &= \text{MRR}_{\text{suf}} - \text{MRR}_{\text{sym}}
\label{eq:delta-mrr-suf}
\end{align}
In both cases, averaging is performed first over hard answers within each query $\mathcal{Q}$, and then over all queries of type $t$. Note that all $\Delta$MRR values reported in Table~\ref{tab:combined_eval} are multiplied by 100 and expressed as percentage points.

\subsection{Hardware}
\label{sec:hardware}

All experiments are conducted on a machine equipped with an NVIDIA H100-20C GPU (20GB VRAM), an Intel Xeon Platinum 8462Y+ CPU (16 cores), and 128GiB RAM. It is worth noting that running CQD-SHAP and computing Shapley values for a given query–answer pair does not require this level of resources and can be performed on any machine capable of loading the link prediction model provided in the CQD work.

\section{Case Study}
\label{sec:case-study}

We present two case studies to qualitatively validate the explanations produced by CQD-SHAP. The first case (Section~\ref{sec:case1}) illustrates a $2i$ query where the Shapley values correctly identify which atom benefits from neural versus symbolic execution, and further demonstrates how CQD-SHAP can be used to explain false answers. The second case (Section~\ref{sec:case2}) demonstrates a more subtle finding: the most important atom identified by CQD-SHAP is not always the one corresponding to the missing link, as illustrated on a $2p$ query.

\subsection{Case 1}
\label{sec:case1}

\begin{figure}[t]
    \centering
    \includegraphics[width=0.75\linewidth]{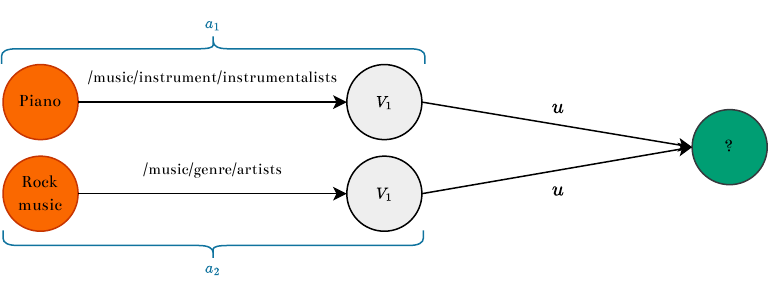}
    \caption{An example of a $2i$ query in the FB15k-237 test data. The query asks for piano instrumentalists who are also artists in the Rock music genre.
    This query has 22 hard answers (e.g., \texttt{Serj Tankian}, \texttt{Jon Bon Jovi}) and 113 easy answers.
    }
    \label{fig:2i}
\end{figure}

By investigating the insights provided by CQD-SHAP explanations for certain query types, we aim to validate their effectiveness. An example of a $2i$ query involving the intersection of two projections is shown in Fig.~\ref{fig:2i}. Suppose we seek to explain \texttt{Paul Weller} as a hard answer to this query. When the query is executed by CQD, this answer has a rank of 61 (after excluding all other answers according to Equation~\ref{eq:filteredlist}). The Shapley values for the two atoms in this query, with respect to this particular answer, are +123.5 and -128.5. This indicates that leveraging neural inference for the first atom has a positive impact on improving the rank, while the effect is the opposite for the second atom.

This can be validated by examining the existing links in the graph. The link (\texttt{Rock music}, \texttt{/music/genre/artists}, \texttt{Paul Weller}) already exists in the graph, so neural inference is unnecessary. Although the CQD link predictor can still assign a high likelihood to this answer, the existence of many possible answers for this link causes the rank to worsen as compared to symbolic reasoning. In contrast, the link (\texttt{Piano}, \texttt{/music/instrument/instrumentalists}, \texttt{Paul Weller}) does not exist in the graph, so leveraging CQD enables the model to predict this answer at a better rank.

If the query is executed in a way that aligns with the insights from the Shapley values---running the first atom neurally and the second atom symbolically---the rank of the target answer improves to 33. However, it should be noted that while such a hybrid execution can lead to an improved ranking of the target entity, this is not guaranteed, since Shapley values represent the \textit{average} marginal contribution of each atom across all possible coalitions, rather than its contribution under a single fixed execution configuration.

Furthermore, according to Equation~\eqref{eq:efficiency}, summing these Shapley values yields $123.5 + (-128.5) = -5$, which exactly matches the difference in ranking obtained by subtracting the neurosymbolic result (rank 61) from the symbolic result (rank 56).
In conclusion, this demonstrates that, for this particular answer, executing the query entirely with the neurosymbolic approach results in a worse ranking than symbolic reasoning, and the Shapley values clearly indicate which parts of the query are responsible and by how much.

It is also worth mentioning that this approach can be used to explain false answers. In the same example, CQD predicts \texttt{Billy Joel} at rank 1, even though this is not a correct answer (neither easy nor hard). For this case, the Shapley values are +219.0 and +39.0, indicating that the first atom contributed most to this incorrect prediction. Such explanations can help users identify and debug cases where the link predictor is underperforming or returning undesirable results.

\subsection{Case 2}
\label{sec:case2}

\begin{figure}[t]
    \centering
    \includegraphics[width=1.0\linewidth]{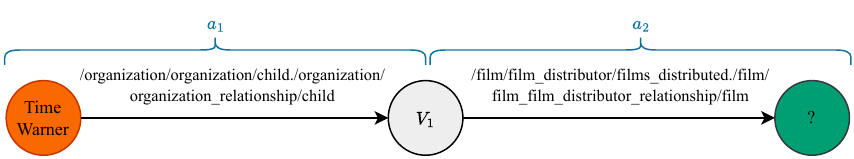}
    \caption{An example of a $2p$ query in the FB15k-237+H test data. The query asks for films which were distributed by companies that are subsidiaries of \textit{Time Warner}.
    }
    \label{fig:2p-appendix}
\end{figure}

In this section, we elaborate on the case study discussed in Section~\ref{sec:discussion}. The query is of type $2p$, asking \textit{``Which films were distributed by companies that are subsidiaries of Time Warner?''}, with anchor entity \textit{Time Warner}, as illustrated in Fig.~\ref{fig:2p-appendix}. The query has 22 hard answers and 292 easy answers, resulting in 314 total answers. The target hard answer of interest is \textit{The Animatrix}.

The missing link in the KG is the \texttt{child} relation between \textit{Time Warner} and \textit{Warner Home Video}, i.e., the first atom. The second atom, connecting \textit{Warner Home Video} to \textit{The Animatrix}, already exists in the training graph. Under purely symbolic execution, \textit{The Animatrix} is ranked at position 9,811, since without the missing link, \textit{Warner Home Video} cannot be reached and no path to \textit{The Animatrix} exists. Under fully neural execution (CQD), the rank improves dramatically to 14.

The CQD-SHAP importance scores are $\phi_{a_1} = -166.0$ and $\phi_{a_2} = +9,963.0$ for the first and second atoms respectively. The second atom is thus identified as the most important, which may seem counterintuitive given that the missing link corresponds to the first atom. This behavior can be explained by examining the structure of the KG more carefully.

The training graph contains 14 known subsidiaries of \textit{Time Warner}, such as \textit{HBO}, \textit{Warner Bros. Entertainment}, \textit{New Line Cinema}, and \textit{Warner Music Group}, with \textit{Warner Home Video} being the additional link present only in the test graph. If we look at the link prediction results for the first atom, all top-10 predictions coincide with known training edges, with \textit{HBO} ranked first. Since these subsidiaries belong to the same corporate family as \textit{Warner Home Video}, they occupy nearby regions of the embedding space, meaning the first atom is already well-covered by symbolic execution alone. As a result, executing the first atom neurally introduces only marginal benefit---and in some cases even slight harm (hence the small negative score of $-166.0$), as the neural model may elevate spurious candidates above \textit{Warner Home Video}.

The second atom, by contrast, receives a large positive score ($\phi_{a_2} = +9{,}963.0$) because none of these top-ranked subsidiaries has a training edge to \textit{The Animatrix}. Under purely symbolic execution of the second atom, all answers unreachable via the known subsidiaries receive a near-zero score, and \textit{The Animatrix} is no exception. It would thus be ranked far down the list alongside the vast majority of candidate entities. The neural model therefore plays a decisive role in this query, elevating the score of \textit{The Animatrix} substantially above this near-zero baseline and improving its rank accordingly.

\section{Runtime Analysis}
\label{sec:runtime}

The runtime for computing Shapley values for all atoms of a query with respect to an answer is shown in Table~\ref{tab:runtime}. Although even the most time-consuming computation takes under 300 milliseconds, the runtime on NELL datasets is consistently higher than on the Freebase datasets, as NELL is a larger KG.

\begin{table}[h]
\caption{Average runtime (in milliseconds) per query type for computing Shapley values of a given query–answer pair on the FB15k‑237 and NELL995 datasets and their harder variants (denoted by +H).}
\label{tab:runtime}
\centering
\scriptsize
\setlength{\tabcolsep}{4.3pt}
\begin{tabular}{@{}lrrrrrrrrrr@{}}
\toprule
\textbf{Dataset} & \textbf{2i} & \textbf{2p} & \textbf{2u} & \textbf{3i} & \textbf{3p} & \textbf{4i} & \textbf{4p} & \textbf{ip} & \textbf{pi} & \textbf{up} \\
\midrule
FB15k-237       & 11.18 & 12.67 & 12.79 &  28.43 &  34.15 &     -- &      -- &  35.72 &  35.28 &  32.89 \\
FB15k-237+H     & 13.77 & 14.95 & 16.55 &  36.51 &  45.19 & 113.16 &  126.74 &  41.04 &  37.84 &  43.29 \\
NELL995         & 36.55 & 41.84 & 37.98 &  93.38 & 110.20 &     -- &      -- & 115.66 & 113.43 & 111.66 \\
NELL995+H       & 44.24 & 43.14 & 49.92 & 102.70 & 100.30 & 256.11 &  294.24 & 105.16 & 107.22 & 108.93 \\
\bottomrule
\end{tabular}
\end{table}

\end{document}